\documentclass[letterpaper, 10 pt, conference]{ieeeconf}  

\IEEEoverridecommandlockouts                              

\usepackage{booktabs}
\usepackage{multirow}
\usepackage{epsfig}
\usepackage{amsmath,bm}
\usepackage{graphicx}
\usepackage{fancyhdr}

\title{\LARGE \bf
STA-VPR: Spatio-temporal Alignment for Visual Place Recognition
}

\author{Feng Lu, Baifan Chen, Xiang-Dong Zhou and Dezhen Song
\thanks{This work was supported by National Key R\&D Program of China (2018YFB1201602), National Natural Science Foundation of China (61976224), National Natural Science Foundation of China (61802361). \emph{(corresponding author: Baifan Chen)}}
\thanks{F. Lu and X. Zhou are with the Chongqing Institute of Green and Intelligent Technology, Chinese Academy of Sciences, Chongqing 400714, China, and also with the University of Chinese Academy of Sciences, China.
        {\tt\small lufengrv@gmail.com; zhouxiangdong@cigit.ac.cn}.}%
\thanks{B. Chen is with the School of Automation, Central South University, Changsha 410083, China.
        {\tt\small chenbaifan@csu.edu.cn.}}%
\thanks{D. Song is with the CSE Department, Texas A\&M University, College Station, TX 77843, USA.
	{\tt\small dzsong@cs.tamu.edu.}}%
}

\begin{document}

\maketitle
\thispagestyle{fancy}
\pagestyle{plain}

\begin{abstract}

	Recently, the methods based on Convolutional Neural Networks (CNNs) have gained popularity in the field of visual place recognition (VPR). In particular, the features from the middle layers of CNNs are more robust to drastic appearance changes than handcrafted features and high-layer features. Unfortunately, the holistic mid-layer features lack robustness to large viewpoint changes. Here we split the holistic mid-layer features into local features, and propose an adaptive dynamic time warping (DTW) algorithm to align local features from the spatial domain while measuring the distance between two images. This realizes viewpoint-invariant and condition-invariant place recognition. Meanwhile, a local matching DTW (LM-DTW) algorithm is applied to perform image sequence matching based on temporal alignment, which achieves further improvements and ensures linear time complexity. We perform extensive experiments on five representative VPR datasets. The results show that the proposed method significantly improves the CNN-based methods. Moreover, our method outperforms several state-of-the-art methods while maintaining good run-time performance. This work provides a novel way to boost the performance of CNN methods without any re-training for VPR. The code is available at https://github.com/Lu-Feng/STA-VPR.

\end{abstract}

\section{Introduction}
Visual place recognition (VPR) is a fundamental task in the navigation and localization of mobile robots, which can be understood as a content-based image retrieval task. The goal of VPR is to determine whether the visual information currently sampled by the robot is from a previously visited place, and if so, which one. However, VPR is a challenging problem because intelligent robots are supposed to operate autonomously over long periods of time in a changing environment. On the one hand, images taken at a specific place can change dramatically over time due to the variations in conditions such as light, weather, and season, as well as perspective difference. On the other hand, multiple different places in an environment can exhibit high similarity, which leads to a problem known as perceptual aliasing \cite{survey}. S{\"u}nderhauf et al. \cite{landmarks} point out that a truly robust VPR system must satisfy three requirements: condition invariance; viewpoint invariance; and generality (without environment-specific training for VPR). However, it is quite difficult to satisfy both viewpoint invariance and condition invariance.
\begin{figure}[!t]
	\centering
	\includegraphics[width=0.88\linewidth]{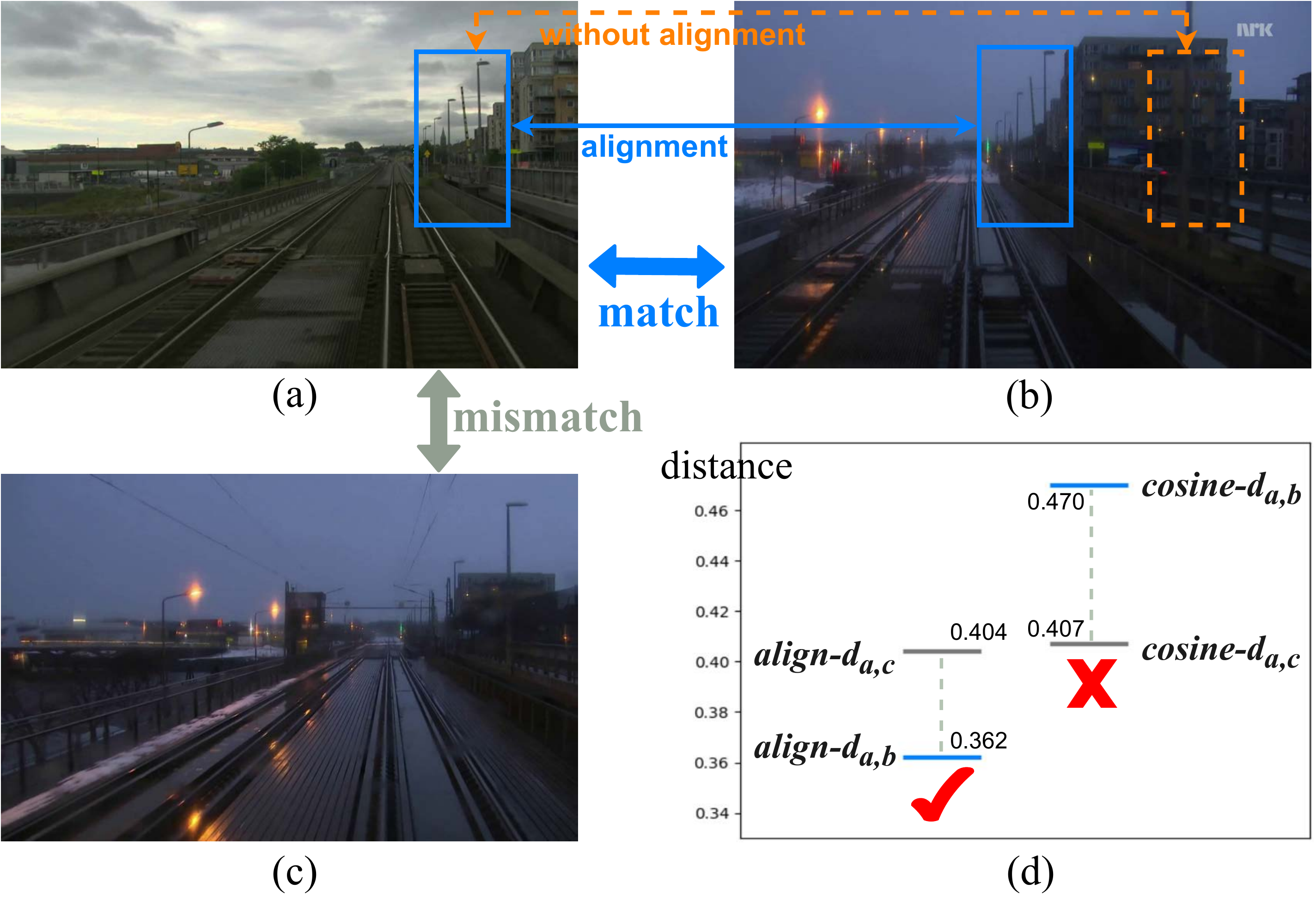}
	\vspace{-0.2cm}
	\caption{(a) and (b) are captured at the same place with different viewpoints. (c) is taken from a different place, but looks similar to (a). (d) shows the image distances between the mid-layer features calculated using two methods (``cosine": cosine distances; ``align": the distances with alignment).
	}
	\vspace{-0.6cm}
	\label{introduction_fig}
\end{figure}

Previous work \cite{sunderhaufIROS2015} has shown that the holistic mid-layer features from Convolutional Neural Networks (CNNs) are highly robust to the appearance (i.e. condition) changes, but they cannot deal with severe viewpoint variations. Meanwhile, Some VPR methods use the bag-of-words \cite{BoW, fab08} or VLAD-related \cite{netvlad, VLAD1, VLAD2} algorithms to aggregate local (or regional) features. Such methods can provide compact image representations and are strongly robust to viewpoint changes. However, they ignore the spatial relationship between local features, and thus tend to suffer from perceptual aliasing. To address these problems, we propose a method to automatically align the local mid-layer features.

When we employ the holistic mid-layer features to represent the image, the spatial position of the features response roughly corresponds to the original image. So, if two images sampled at the same place exhibit severe viewpoint changes, measuring the distance without alignments may falsely increase the measurement (see Fig. \ref{introduction_fig}). This is why the most VPR methods using mid-layer features lack robustness to significant viewpoint changes. And the correct result will be obtained after alignment. Specifically, we split the holistic feature from the middle layer of CNNs into multiple local features, then align them using an adaptive dynamic time warping (DTW) algorithm while measuring the distance between images. In this way, each local feature of the query image can be compared with the feature representation of the semantically corresponding part in the matched image. Hence, the algorithm gets the correct distance measurement and improves the robustness of the mid-layer features against viewpoint changes. 

Meanwhile, the single-image-based VPR method ignores that the robotic vision perception is continuous in space and time, then some methods \cite{seqslam, SMART, Sequence, Naseer, hmm, CNNSeqSLAM} using image sequence are proposed. Follow this line, we also align the images from the temporal domain, i.e., utilize image sequence for VPR. The local matching DTW (LM-DTW) algorithm is used to perform image sequence matching, which is actually proposed in our previous work \cite{ours}. Since most sequence-based place recognition methods use global descriptors to represent images, which are condition-invariant but not viewpoint-invariant. The spatial alignment idea can also be used to improve the performance of other existing sequence matching approaches such as SeqSLAM \cite{seqslam}.

We name our complete method as STA-VPR (spatio-temporal alignment for visual place recognition), and the main contributions of this letter are:

\textbf{1)} We analyze the reason why the mid-layer deep features lack robustness to large viewpoint changes, then propose the idea of aligning CNN-based local features to address it.

\textbf{2)} We propose an adaptive DTW algorithm to align the local features obtained by splitting the holistic mid-layer features while calculating the image distance. Then we combine it with our previous work to form the STA-VPR method.

\textbf{3)} The testing results on five challenging datasets show that STA-VPR can be used as a wrapper method to boost the performance of common CNN models. Besides, our method outperforms several state-of-the-art methods, showing strong robustness to both severe appearance and viewpoint changes.

\section{Related Work}
Our proposed method is based on DTW, so we briefly review the works related to VPR and DTW in this section.

\emph{Visual Place Recognition:}
VPR is an active research topic in both the robotics and computer vision communities. Traditional VPR methods, such as FAB-MAP \cite{fab08}, typically follow computer vision approaches to encode handcrafted features like SURF \cite{SURF} into the bag-of-words models. These methods can achieve large-scale VPR without significant appearance changes but have difficulty dealing with more realistic changing environments. Some approaches such as SeqSLAM \cite{seqslam}, SMART \cite{SMART} and the HMM-based method \cite{hmm} leverage image sequence matching to achieve robust VPR under dramatic variations in illumination, weather and season. However, such methods often fail to cope with severe viewpoint changes. 

With the great success of CNNs on a broad variety of computer vision tasks, the focus of VPR has switched to using the CNNs to extract more general deep features. Some of these works \cite{landmarks, sunderhaufIROS2015, categorization1, categorization2, landmarks2, SSM1, SSM2} use CNN models pre-trained on IamgeNet \cite{ImageNet} or Places365 \cite{places} dataset that is built for object or scene classification task (rather than VPR). Yet others \cite{netvlad, landmarks3, MetricLearning, SPED, semantic} require environment-specific training on VPR-related datasets such as SPED \cite{SPED}. Since mid-layer features lack robustness to severe viewpoint changes, some methods \cite{categorization1, categorization2} leverage the semantic information from the higher-order layer of place categorization networks to improve the robustness against viewpoint changes, while some other methods \cite{landmarks, landmarks2, landmarks3} focus on mining discriminative landmarks for VPR. Recently, some works have been devoted to the use of lightweight \cite{VLAD2} or compact \cite{compact} CNN architecture for VPR. Meanwhile, some methods use multi-feature fusion to improve performance. For example, Hausler et al. \cite{fusion1} use the features fused by multiple image processing methods. Chen et al. \cite{fusion2} utilize an attention fusion model to merge context information extracted from different layers of CNN. However, there are few works that focus on aligning local features with spatial constraints for VPR. In contrast to these approaches, our method only uses local mid-layer features without mining landmarks, and aligns them to achieve viewpoint-invariant and condition-invariant VPR.

\emph{Dynamic Time Warping: }
DTW is a well-known algorithm for measuring the distance between two time sequences that may vary in time or speed, while also yielding the optimal alignment between these two sequences. It was first widely used in speech recognition \cite{DTW} to report the probability of two speech sequences expressing the same spoken phrase. Thereafter, it was successfully applied in some other fields (e.g. data mining \cite{DTW2000KDD}) and became a classical distance metric in time sequence analysis. Besides, there are also some improved versions of DTW such as shapeDTW \cite{shapeDTW}. Related to computer vision applications, Chen et al. \cite{imageDTW} use DTW to align two image sequences taken by the left and right cameras. And Deng et al. \cite{videoDTW} utilize DTW to measure the similarity between two videos, successfully solving the matching difficulty caused by different time scales. To our best knowledge, this paper is among the first to apply DTW to align the local features of the image.

\section{STA-VPR}
This section describes the STA-VPR method in detail. We first introduce the place description. Then, we present the adaptive DTW algorithm to align local features for image distance measurement, as well as use Gaussian Random Projection and restricted alignment for fast recognition. Finally, we apply LM-DTW to align images for sequence matching.

\begin{figure*}[!t]
	\centering
	\includegraphics[width=0.8\linewidth]{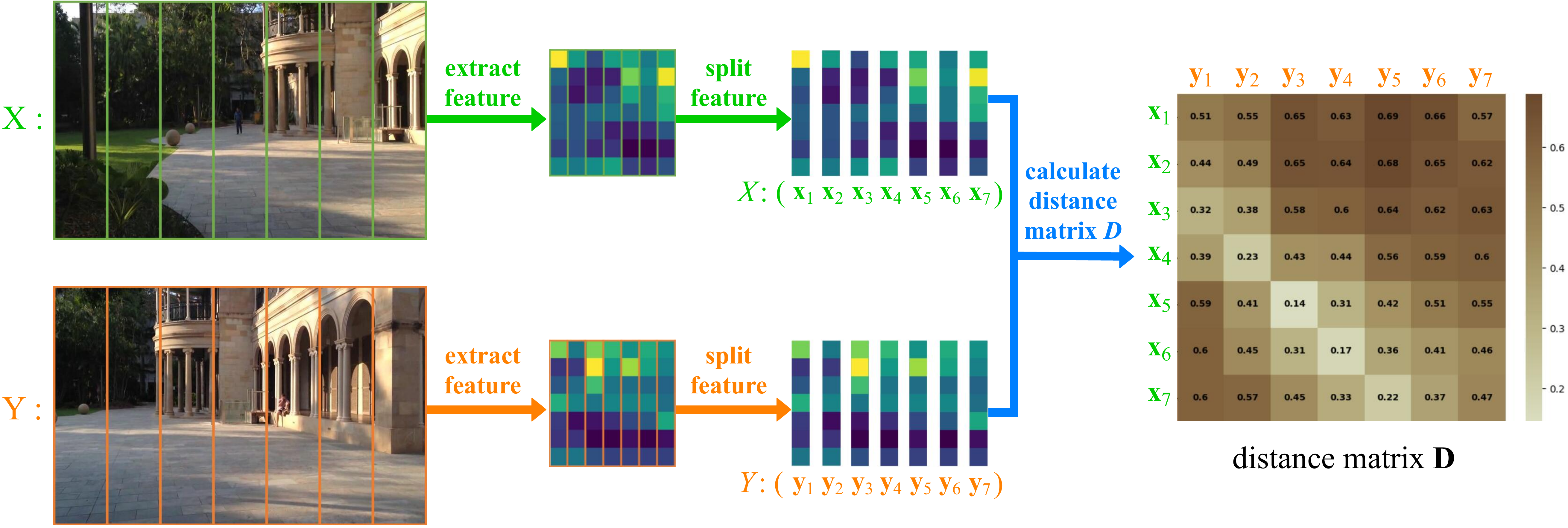}
	\vspace{-0.1cm}
	\caption{
		The process of computing the distance matrix $\mathbf D$ of two images (i.e. feature sequences). See the text for details.
	}
	\vspace{-0.5cm}
	\label{spatial_alignment}
\end{figure*}
\subsection{Place Description}
Given an image, we use a CNN model to extract the mid-layer feature map, which is a $W$$\times$$H$$\times$$C$-dimensional (width by height by channel) tensor. Then we split it vertically into $W$ separate $H$$\times$$C$-dimensional matrixes and flatten each matrix to a vector. In this way, a place image can be represented as $W$ local features, each of which is an $H$$\times$$C$-D vector. For example, we use the DenseNet161 \cite{Densenet} pre-trained on Places365 to extract deep features from the middle layer (the \emph{relu1} layer of \emph{denselayer10} in \emph{denseblock4}). The feature is 7$\times$7$\times$1488-D, so a place image can be represented as 7 local features (10416-D vectors). As a note, the local features obtained by splitting the holistic features are similar to the global descriptors of image segments, which has the potential to exhibit both viewpoint invariance of local features and condition invariance of holistic image descriptors \cite{survey}.

\subsection{Spatial Alignment for Image Distance Measurement}
\label{sec_SpatialAlignment}
Although the local feature can be robust against  viewpoint changes, measuring the distance between the local features of two images without alignments may lead to inappropriate measurement. To address this, we consider the $W$ ($W$=7 in this example) local features as a spatial sequence and propose an adaptive DTW algorithm to automatically align the two sequences, while computing the distance of them.

As illustrated in Fig. \ref{spatial_alignment}, suppose there are image X and image Y (X for query and Y for reference), which can be represented as feature sequence $X$ ($\mathbf x_1,..., \mathbf x_i,..., \mathbf x_W$) and $Y$ ($\mathbf y_1,..., \mathbf y_j,..., \mathbf y_W$) respectively. We utilize the cosine distance to measure the distance of local features, which is defined as
\vspace{-0.05cm}
\begin{equation}
	\vspace{-0.05cm}
	\label{eq_1}
	d_{i,j} = 1-\frac{\mathbf x_i\cdot\mathbf y_j}{\|\mathbf x_i\|\times\|\mathbf y_j\|}\quad i,j \in \{1,2,...,W\},
\end{equation}
where $d_{i,j}$ is the distance between the $i$-th feature $\mathbf x_i$ in the sequence $X$ and the $j$-th feature $\mathbf y_j$ in the sequence $Y$. The values of features are non-negative thanks to $relu$ activation, so the range of cosine distance is [0,1]. And the smaller the distance, the more similar the features. Then, a distance matrix is established by the cosine distances between pairwise features in the two sequences. So we can get a $W\times W$ distance matrix $\mathbf D$ between the sequence $X$ and $Y$, whose $(i, j)$-element $d_{i,j}$ is calculated as Eq. \eqref{eq_1}.

The goal of the adaptive DTW algorithm is to find an optimal warping path in matrix $\mathbf D$ that minimizes the total (weighted) distance. Meanwhile, the alignments between the sequence $X$ and $Y$ is indicated by the points in this warping path $P$:
\vspace{-0.1cm}
\begin{equation}
	\vspace{-0.1cm}
	P=\{p_1,p_2,...,p_k,...,p_K\},
\end{equation}
where $\max(W,W)\leq K < W+W-1.$

The path is constrained to satisfy several conditions:

\textbf{ 1) Boundary:} $p_1 = (1, 1)$ and $p_K = (W,W)$.

\textbf{ 2) Continuity:} If $p_k = (d, e)$ and $p_{k+1} = (d', e')$, then $d'-d \leq 1$ and $e'-e \leq 1$. This implies that skipping an element in sequences to match is invalid, ensuring that every local feature will be used for distance measurement.

\textbf{ 3) Monotonicity:} If $p_k = (d, e)$ and $p_{k+1} =(d', e')$, then $d'-d \geq 0$ and $e'-e \geq 0$. This limits the points in the warping path must be monotonous in spatial order.

According to the continuity and monotonicity conditions,  if the warping path has already passed through the point $(i,j)$, the next point must be one of the following three cases: $(i+1,j)$, $(i,j+1)$, and $(i+1,j+1)$.

\begin{figure*}[!t]
	\centering
	\includegraphics[width=0.88\linewidth]{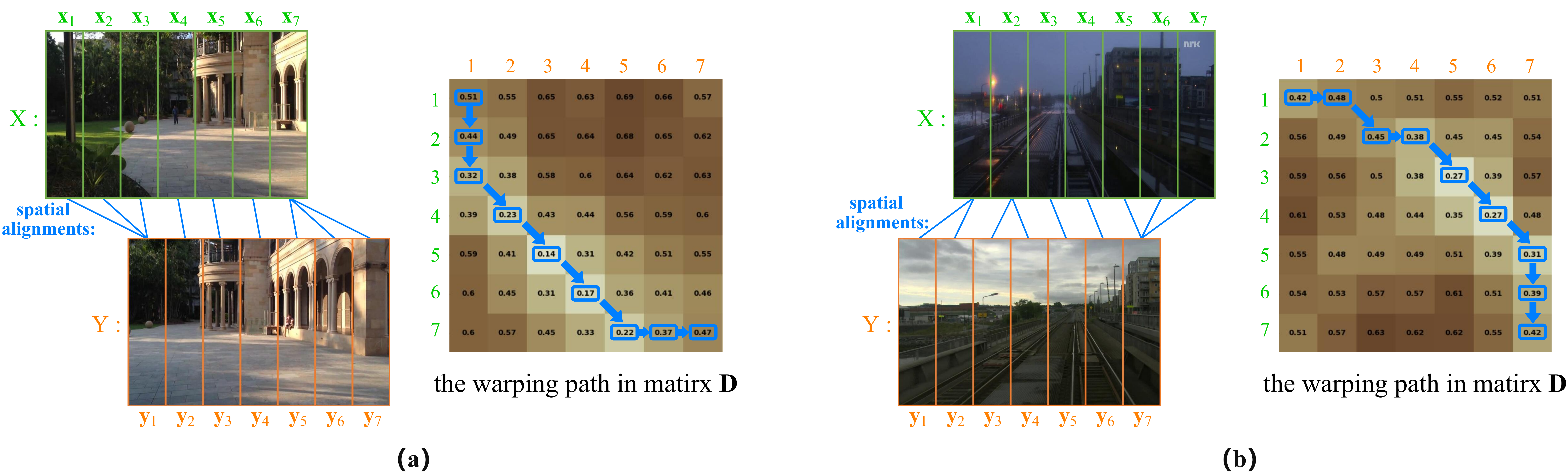}
	\vspace{-0.2cm}
	\caption{
		(a) and (b) are two different instances of the spatial alignment. Each pair of images is sampled at the same place with different viewpoints. The correct alignments between the image segments (or local features) are obtained according to the warping path.
	}
	\vspace{-0.5cm}
	\label{spatial_alignment_sample}
\end{figure*}

To get this optimal warping path, we use dynamic
programming to yield a $W\times W$ cumulative distance matrix $\mathbf S$, whose $(i, j)$-element $s_{i,j}$ is the cumulative distance of the optimal path from (1,1) to ($i,j$). The $s_{i,j}$ is calculated as
\begin{equation}
	\label{eq_3}
	\resizebox{.91\hsize}{!}{$
		s_{i,j}=\begin{cases}
		d_{i,j} & i=1,  j=1 \\
		d_{i,j}+s_{i,j-1} & i=1,j\in\{2,...,W\} \\
		d_{i,j}+s_{i-1,j} & i\in\{2,...,W\},j=1 \\
		\min \{a\cdot d_{i,j}+s_{i-1,j-1} \text{, } d_{i,j}+s_{i-1,j} \text{, } d_{i,j}+s_{i,j-1} \} & i,\ j\in \{2,...,W\}
		\end{cases},
		$}
\end{equation}
where the adaptive parameter $a$ reflects the degree of viewpoint changes. And it is defined as
\vspace{-0.1cm}
\begin{equation}
	\vspace{-0.1cm}
	\label{eq_4}
	\resizebox{.91\hsize}{!}{$
		a=\sqrt{1+\sigma |I_{\lceil W/2 \rceil}-{\lceil W/2 \rceil}|} \quad (W=7\Rightarrow {\lceil W/2 \rceil}=4),
		$}
\end{equation}
where $\sigma$ is a constant (default setting is 1) and $I_{\lceil W/2 \rceil}$ is the index of the feature in sequence $Y$ that is most similar to central  feature $\bm x_{\lceil W/2 \rceil}$ of sequence $X$, so it is computed as
\vspace{-0.1cm}
\begin{equation}
	\vspace{-0.1cm}
	I_{\lceil W/2 \rceil}=\mathop{\arg\min}_{j}\,{d_{\lceil W/2 \rceil,j}} \qquad j \in \{1,2,...,W\}.
\end{equation}

The two sequences are aligned from the start point (1, 1) to the end point ($W$, $W$). As a result, the alignments of them is indicated by the warping path. Fig. \ref{spatial_alignment_sample} shows two instances of spatial alignment. Take Fig. \ref{spatial_alignment_sample} (a) for example, the spatial alignments between all semantically corresponding local features (such as $\mathbf x_3$ and $\mathbf y_1$) are included in the path. However, there are also a few alignments between non-corresponding features like $\mathbf x_1$ and $\mathbf y_1$. This is due to the changes in viewpoint, the image X contains something that does not exist in the image Y. Preserving these non-corresponding alignments is necessary to maintain the spatial order of local features and measure image distance accurately.

Considering that the number of elements in the warping path is uncertain, we use the total cost $C_{W,W}$ of the path to normalize the cumulative distance $s_{W,W}$, and finally get the distance between image X and Y as
\begin{equation}
	d'_{X,Y} =\frac{s_{W,W}}{C_{W,W}} \qquad (C_{W,W}=\sum\nolimits_{k=1}^Kc_k),
\end{equation}
where the path cost $c_k$ is decided by 
\begin{equation}
	\resizebox{.91\hsize}{!}{$
		c_k=\begin{cases}	
		a & a\cdot d_{i,j}+s_{i-1,j-1} < \min(d_{i,j}+s_{i-1,j} , d_{i,j}+s_{i,j-1}) \\
		1 & else 
		\end{cases}.
		$}
\end{equation}
After normalization, the adverse effect of non-corresponding alignments is reduced and the image distance ranges in $[0,1]$.

It should be noted that, in addition to the normalization details, the adaptive DTW differs from vanilla DTW only in the fourth term of Eq. \eqref{eq_3}. When there is no or only slight changes in viewpoint, we will get $a=1$ (because $I_{\lceil W/2 \rceil}={\lceil W/2 \rceil}=4$), and the adaptive DTW will degenerate to vanilla DTW, namely, the fourth term of Eq. \eqref{eq_3} becomes
\begin{equation}
	\resizebox{.91\hsize}{!}{$
		s_{i,j}=d_{i,j}+\min \{ s_{i-1,j-1},\: s_{i-1,j},\: s_{i,j-1} \} \quad i,\ j\in \{2,...,W\}.
		$}
\end{equation}

Additionally, our method aligns local features only in the horizontal dimension of space. This is based on the fact that the height of the camera on a mobile robot is fixed, so the viewpoint varies mainly in the horizontal direction. For special cases (e.g. the VPR for UAV) that violate it, we can further split the local features, aligning them first in the vertical and then in the horizontal direction (it may be time-consuming).

\subsection{Dimensionality Reduction and Restricted Alignment}
The process of computing the distance matrix $\mathbf D$ includes 49 distance calculations, while the dimension of the feature vector is large, making the matching process expensive. In some experiments, we will apply Gaussian Random Projection (GRP) \cite{GRP, landmarks} to reduce the original dimensions of each local feature to 512 dimensions. Additionally, we also propose to use restricted alignment (RA) to reduce the number of distance calculations. As shown in Fig. \ref{spatial_alignment_sample}, although there are severe viewpoint changes between the two images, the warping path still does not pass through the lower-left and upper-right areas of the distance matrix $\mathbf{D}$. So we can directly set the points in these areas to $+\infty$. More specifically,
\begin{equation}
	\resizebox{.91\hsize}{!}{$
		d_{i,j} = \begin{cases}
		1-\frac{\mathbf x_i\cdot\mathbf y_j}{\|\mathbf x_i\|\times\|\mathbf y_j\|}\ & |i-j|<\xi \text{ and } i,j \in \{1,2,...,W\}\\
		+\infty & |i-j|\ge \xi \text{ and } i,j \in \{1,2,...,W\}
		\end{cases},
		$}
\end{equation}
where we set the parameter $\xi=3$ when $W=7$. To verify the original performance of our method, we only perform GRP and RA in partial experiments, not by default.

\subsection{Temporal Alignment for Image Sequence Matching}
\begin{figure}[!t]
	\centering
	\includegraphics[width=0.9\linewidth]{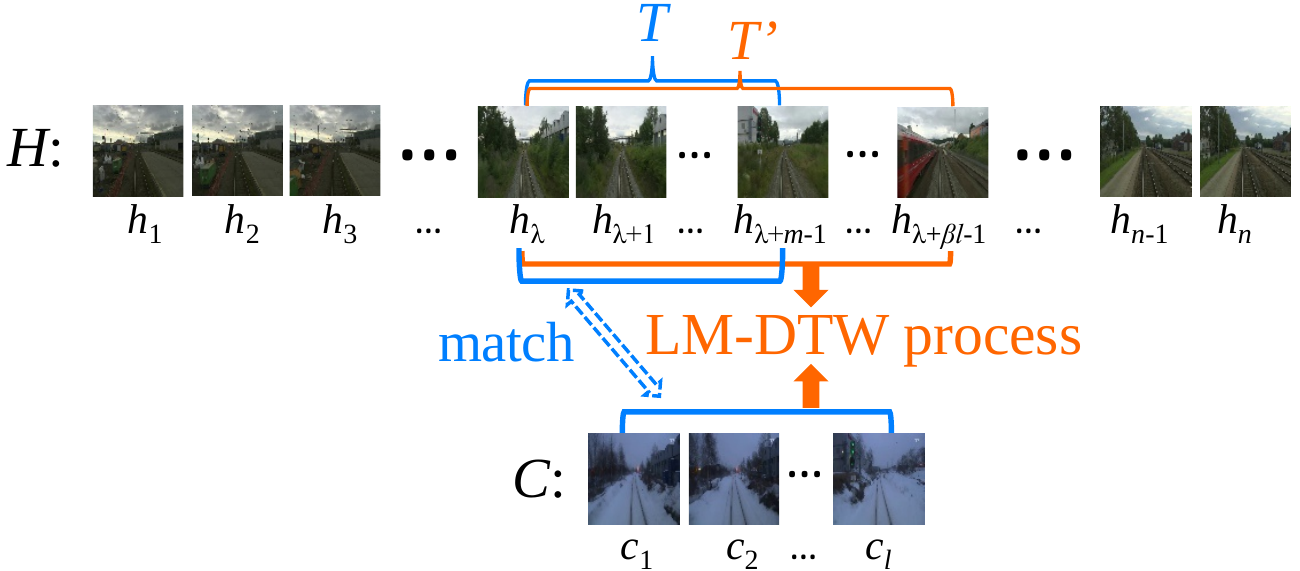}
	\vspace{-0.2cm}
	\caption{The schematic of using LM-DTW for sequence place recognition, i.e. temporal alignment. See the text for details.}
	\vspace{-0.6cm}
	\label{sequence_match}
\end{figure}

In subsection \ref{sec_SpatialAlignment}, we have described the measurement of image distance. Now we will introduce the image sequence distance measurement based on the idea of temporal alignment. The basic algorithm (LM-DTW) is first proposed in our previous work \cite{ours}, in which we call it ``improved DTW". As can be appreciated in Fig. \ref{sequence_match}, suppose that a mobile robot has taken an image sequence $H$ ($h_1, h_2,..., h_n$) during the historical navigation, as well as an image sequence $C$ ($c_1, c_2,..., c_l$) from the current place. What we need to do is find a subsequence $T$ ($h_\lambda, h_{\lambda+1},..., h_{\lambda+m-1}$) from the sequence $H$. This subsequence is the best match for the sequence $C$. And we also need to determine whether they are actually sampled from the same place based on their distance. Note that the first frame $h_\lambda$ and length $m$ of the sequence $T$ are unknown. Meanwhile the $l$  (length of $C$) is a constant that we can set, so the $m$ and $l$ are not necessarily equal. Obviously, we can use the vanilla DTW algorithm to align the time sequence $C$ and a ``candidate $T$" while measuring the sequence distance between them.

There is a straightforward way to locate the true subsequence $T$ in sequence $H$. It is to compute the sequence distances between the sequence $C$ and all ``candidate $T$" (i.e. all subsequences of sequence $H$), then find the one with the smallest distance. In this way, we need to apply a two-layer loop, one layer to control the length and another layer to control the starting frame of the ``candidate $T$". Meanwhile, the DTW process consumes $O(l\times m)$ time in the loop, where $m$ is a variable and $l$ is a constant. Consequently, the total time complexity of this method is $O(n^3)$, which is undesirable.

To achieve a linear time complexity, we propose a local matching DTW (LM-DTW) algorithm. Assuming that the first frame of the sequence $T$ is $h_\lambda$ (see Fig. \ref{sequence_match}), we perform the LM-DTW process on the sequence $C$ and the sequence $T'$ that starts at the frame $h_\lambda$ and is not shorter than $T$ to locate the sequence $T$ in $T'$. The length of the sequence $T'$ is denoted as $k$, so that $m\le k\le n-\lambda+1$. In our method, we assume that $m\le\beta l$ and therefore $k=\beta l$ can be set, where $\beta$ is a constant. 
This assumption is based on a simple prior knowledge that the ratio of the average speed of a mobile robot through a same place twice usually does not exceed a constant $\beta$.

In the LM-DTW algorithm, we first calculate an $l\times k$ distance matrix $\mathbf D'_{CT'}$ between sequence $C$ and $T'$. Then, we utilize Eq. \eqref{eq_3} to yield an $l\times k$ cumulative distance matrix $\mathbf S'_{CT'}$. These steps are similar to subsection \ref{sec_SpatialAlignment}. 
The only difference is that now we directly set the adaptive parameter $a$ equal to 1, instead of using Eq. \eqref{eq_4} to decide. However, the present task is to locate the sequence $T$ from $T'$ (i.e. determine the value of $m$) and measure the distance of the sequence $C$ and $T$. From the dynamic programming of yielding the matrix $\mathbf S'_{CT'}$, we can get that the element $s'_{l,x}$ in the last row of $\mathbf S'_{CT'}$ is the cumulative distance of the optimal warping path from (1,1) to ($l,x$). After normalization, the $\frac{s'_{l,x}}{C'_{l,x}}$ represents the  distance between the sequence $C$ and the first $x$ frames of sequence $T'$. So the $x$ that minimizes $\frac{s'_{l,x}}{C'_{l,x}}$ is equal to the length $m$ of sequence $T$. That is
\vspace{-0.1cm}
\begin{equation}
	\vspace{-0.1cm}
	m=\mathop{\arg\min}_{x}\,(\frac{s'_{l,x}}{C'_{l,x}}).
\end{equation}
Meanwhile, we can get that the distance between the sequence $C$ and $T$ is min($s'_{l,x}/C'_{l,x}$). And their alignment is indicated by the warping path.

In this fashion, we are equivalent to changing the endpoint of the desired warping path (no longer ($l,k$)), or we are relaxing the boundary condition of the vanilla DTW. The LM-DTW process consumes $O(\beta l^2)$ time, i.e., the time complexity is $O(1)$. Now, we only need to determine the starting frame $h_\lambda$ of the sequence $T$ by sliding it $n$ times from $h_1$ to $h_n$ and then finding the one with the minimum sequence distance. Consequently, the total time complexity of our method is $O(n)$. In addition, we can execute it in parallel instead of sliding $n$ times to achieve faster recognition. Overall, we have completed the temporal alignment between the sequence $C$ and the sequence $T$ that may start from any frame in the sequence $H$ and have a length in the range of [$1,\beta l$].

\section{Experiment}
In this section, we first describe the datasets and evaluation metrics. We then study the effects of whether to align and different alignment strategies. Finally, we compare STA-VPR with state-of-the-art methods and show its runtime. To prove that STA-VPR can be regarded as a wrapper method suitable for common CNNs, we also apply it to VGG16 \cite{vgg}.
\subsection{Datasets, Performance Evaluation, and Parameter Setup}
We test the proposed method on four benchmark VPR datasets (Nordland, Gardens Point \cite{sunderhaufIROS2015}, UA \cite{UA}, Berlin\_A100 \cite{landmarks2}) and one synthetic dataset (synth-Nord). The left-right shift is the most common viewpoint change in VPR, especially for intelligent driving. And the viewpoint changes shown in the datasets we used are also mainly in the horizontal direction. Their key information is summarized in Table \ref{tableDataset} and sample images are shown in Fig. \ref{dataset}. The details are as follows:
\begin{figure}[!t]
	\centering
	\includegraphics[width=0.88\linewidth]{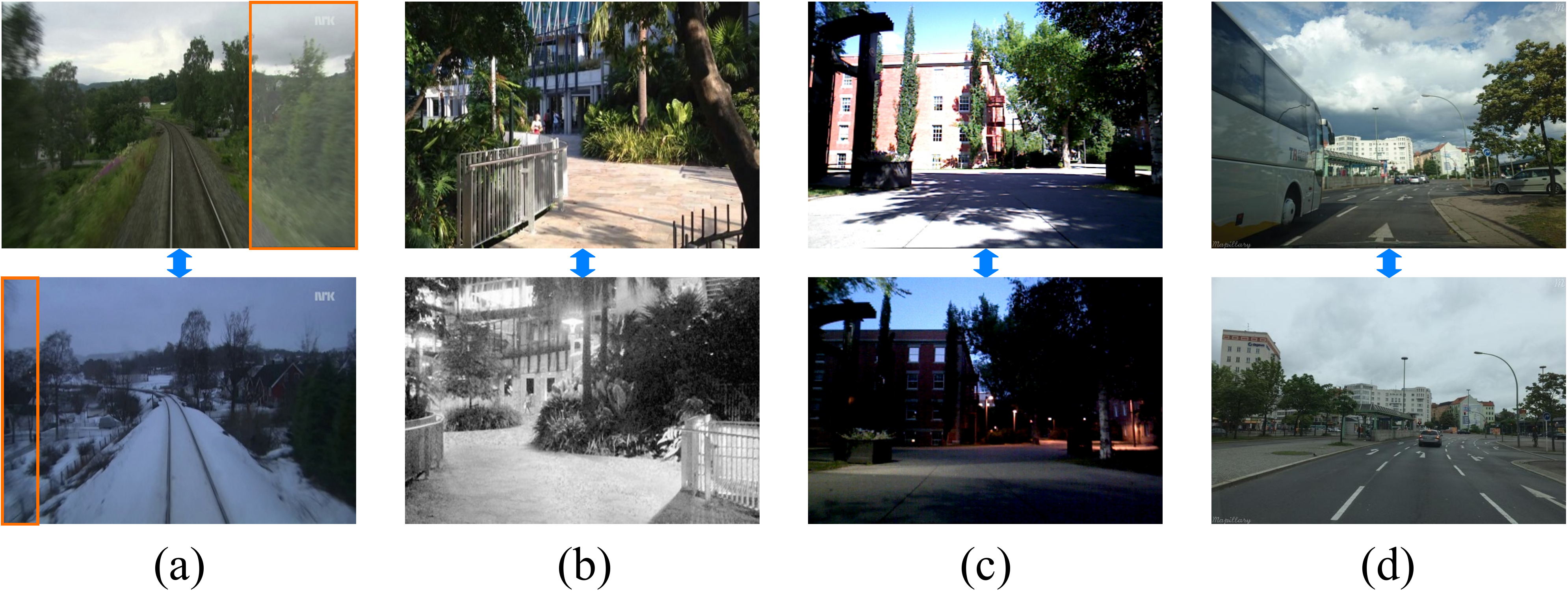}
	\vspace{-0.4cm}
	\caption{
		Sample images from matched location of each experimental dataset (top for reference, bottom for query): (a) Nordland and synth-Nord (after removing the orange box); (b) GP(DL\&NR); (c) UA; (d) Berlin\_A100.
	}
	\vspace{-0.2cm}
	\label{dataset}
\end{figure}
\begin{table}[!t]
	\caption{Summary of the datasets.}
	\vspace{-0.5cm}
	\begin{center}
		\resizebox{.95\hsize}{!}{$
			{\begin{tabular}{|c|c|c|c|c|}
				\hline
				\multirow{2}{*}{\bf Dataset} & \multirow{2}{*}{\begin{tabular}[c]{@{}c@{}}\bf No. of frames \\(reference/query)\end{tabular}} & \multirow{2}{*}{\begin{tabular}[c]{@{}c@{}}\bf Appearance\\ \bf variation\end{tabular}} & \multirow{2}{*}{\begin{tabular}[c]{@{}c@{}}\bf Viewpoint\\ \bf variation\end{tabular}} & \multirow{2}{*}{\begin{tabular}[c]{@{}c@{}}\bf Tolerance\\ (frames)\end{tabular}} \\
				& & & &\\ \hline
				Nordland &3584/3584 &severe &none &$\pm$3\\
				\hline
				synth-Nord &3584/3584 &severe &severe &$\pm$3\\
				\hline
				GP(DL\&NR) &200/200 &severe &medium &$\pm$2\\ 
				\hline
				GP(DL\&DR) &200/200 &minor &medium &$\pm$2\\ 
				\hline
				UA & 646/646 & severe & minor &$\pm$3\\
				\hline
				Berlin\_A100 &85/81 &medium &medium &$\pm$2\\ 
				\hline
				\end{tabular}}
			$}
	\end{center}
	\vspace{-0.8cm}
	\label{tableDataset}
\end{table}

\textbf{The Nordland and Gardens Point (GP) Datasets} are used in our previous work \cite{ours}. But this time the reference and query datasets we used have the same number of images. For Nordland, the summer images are used as reference data, while the winter images are used as query data. For GP, we use day-left (reference) and day-right (query) traverses to form GP(DL\&DR) dataset, as well as use day-left (reference) and night-right (query) traverses to form GP(DL\&NR) dataset. 

\textbf{The synth-Nord dataset} is synthesized using the Nordland dataset. Considering that the Nordland dataset exhibits severe appearance changes but no viewpoint changes, we crop the right 30\% of all summer images and the left 10\% of all winter images to synthesize severe viewpoint changes.

\textbf{The UA dataset} has been collected by a mobile robot on the campus of University of Alberta. We use the day and evening traverses as reference and query data respectively. There are severe condition variations and slight viewpoint changes between two traverses.

\textbf {The Berlin\_A100 dataset} is downloaded from the crowdsourced platform $Mapillary$. It consists of two traverses of a same route, with medium appearance and viewpoint changes as well as some variations due to dynamic vehicles. Besides, there exists a speed change between two traverses.

We evaluate the recognition performance mainly based on F1-score. For each dataset, ground truth is provided by the frame-level correspondence of the dataset. The (ground truth of the) midpoint of the current query sequence is compared with the midpoint of the result sequence. A match is regarded as a true positive only when the sequence distance is less than a threshold and the difference between the recognition result and the ground truth is within a tolerance (see Table \ref{tableDataset}). The results of the first and last $l/2$ frames in the query dataset are compensated with temporal frame alignment, same as \cite{ours}.

We summarize the critical parameters of the STA-VPR method in Table \ref{tableParameter}. These parameters and values are common to the above datasets and are also suitable for use in most other environments. The $\xi$ is related to the degree of reducing the number of distance calculations in spatial alignment. A smaller $\xi$ requires fewer distance calculations, but it is more likely to fail to align local features. we set $\xi=3$ when $W=7$. The $l$ is the length of the sequence $C$, and the temporal alignment usually brings significant improvement when $l\ge10$. We set $l=20$. The $\beta$ is the ratio of $k$ to $l$, i.e. $k=\beta l$. A bigger $\beta$ is more rigorous but more time-consuming, and we set $\beta=2$.

\begin{figure}[!t]
	\centering
	\includegraphics[width=0.75\linewidth]{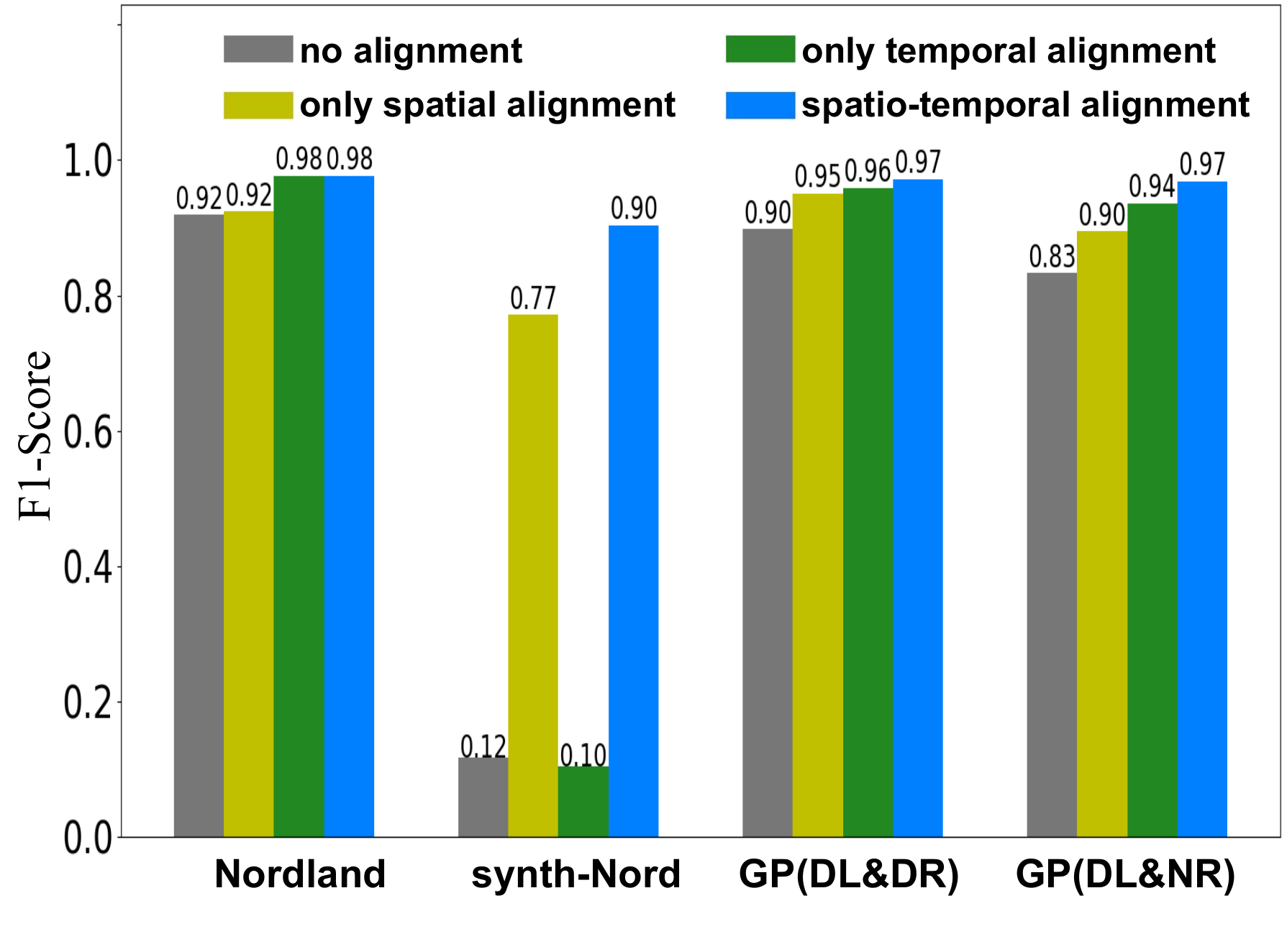}
	\vspace{-0.6cm}
	\caption{
		F1-score comparison of different ablated versions.
	}
	\vspace{-0.2cm}
	\label{alignment_F-score}
\end{figure}

\begin{table}[!t]
	\caption{Parameter List.}
	\vspace{-0.4cm}
	\begin{center}
		\resizebox{.9\hsize}{!}{$
			\begin{tabular}{|c|c|c|c|}
			\hline
			\bf Parameter   & $\xi$ \, $(1\leq\xi\leq W)$  & $l$  \, $(l\ge2)$  & $\beta$ \, $(\beta>1)$  \\ \hline
			\bf Description & parameter in Eq. (9) & length of $C$ & ratio of $k$ to $l$ \\ \hline
			\bf Value       & 3  & 20  & 2  \\ \hline
			\end{tabular}
			$}
	\end{center}
	\vspace{-0.8cm}
	\label{tableParameter}
\end{table}

\subsection{Ablation study (effects of whether to align)}
\label{sec_AblationStudy}
In order to study the effectiveness of the spatial and temporal alignment in our method, a set of ablation experiments is conducted by performing multiple ablated versions of the method:
\begin{itemize}
	\item\textbf {no alignment:} Use cosine distance to measure the distance of holistic features for single image matching.
	\item\textbf {only spatial alignment:} Only use spatial alignment for single image matching.
	\item\textbf {only temporal alignment:} Only use temporal alignment for image sequence matching.
	\item\textbf {spatial-temporal alignment:} Our complete method, i.e., both spatial and temporal alignments are used.
\end{itemize}

The F1-scores of different ablated versions are shown in fig. \ref{alignment_F-score}. The results show that spatial alignment brings a substantial improvement when there exist viewpoint changes. In particular, the no alignment fails completely on the synth-Nord dataset where the viewpoint changes drastically, but a satisfactory result is obtained after spatial alignment. Of course, spatial alignment cannot bring any improvement when there is no viewpoint change (i.e. Nordland).

\begin{figure}[!t]
	\centering
	\includegraphics[width=0.75\linewidth]{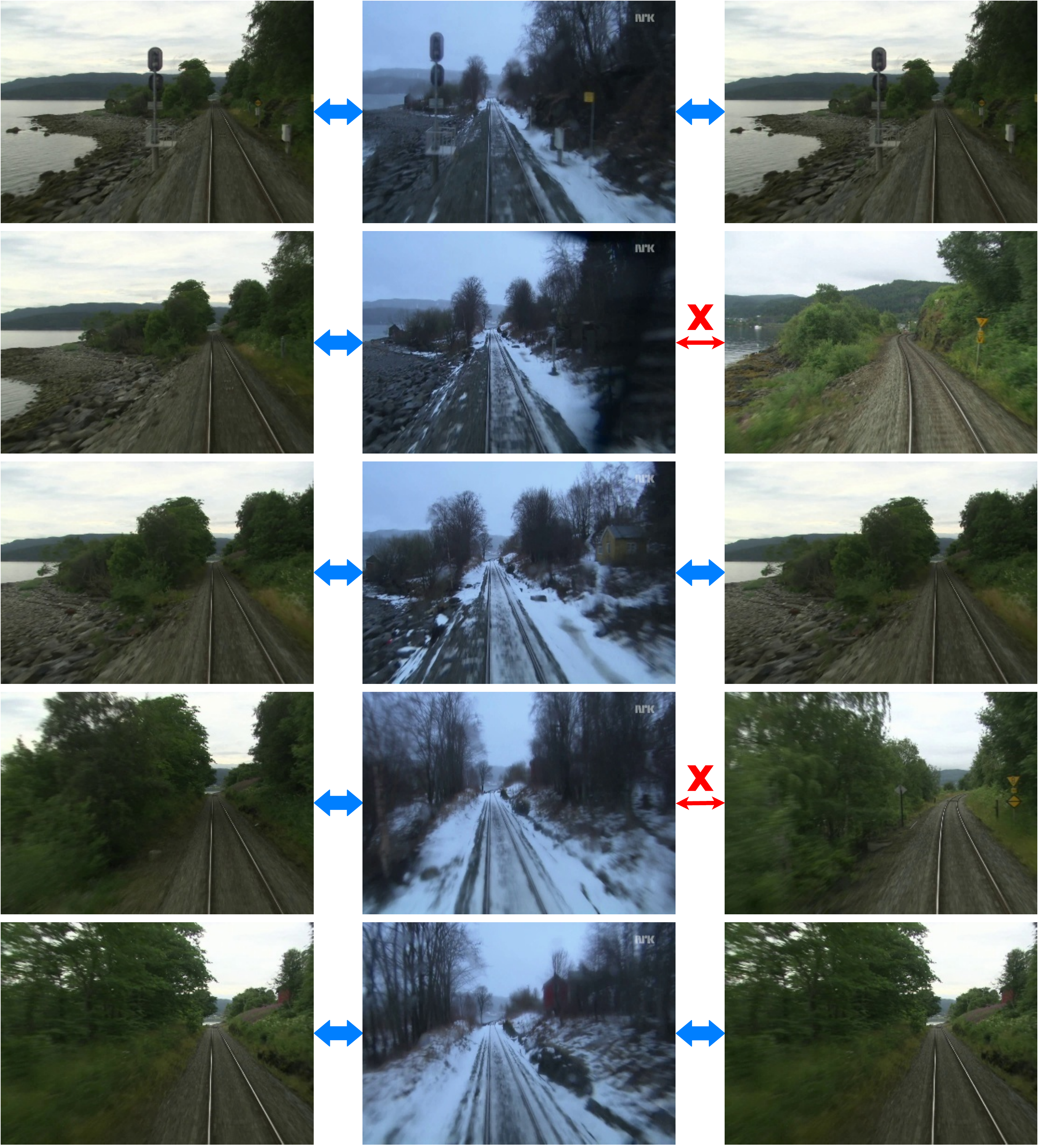}
	\vspace{-0.3cm}
	\caption{
		The middle is the current query sequence $C$, the left is the result of spatio-temporal alignment, the right is the result of single-image-based method with only spatial alignment. The blue arrow indicates the correct match and the red indicates an error.
	}
	\vspace{-0.3cm}
	\label{temporal_alignment_result}
\end{figure}

Meanwhile, the temporal alignment almost always improves performance. One exception is on synth-Nord. Since the no alignment is a complete failure, even if we use the sequence matching method (i.e. only temporal alignment), it doesn't help because most of the images in a sequence are misjudged. However, there are enough images to be correctly matched after spatial alignment, in which case temporal alignment further improves performance. Fig. \ref{temporal_alignment_result} shows an example of temporal alignment, although some images in a sequence are mismatched due to perceptual aliasing, as long as enough images can be correctly matched, the method with temporal alignment will yield the correct result.

\subsection{Effects of different alignment strategies}
\begin{figure}[!t]
	\centering
	\includegraphics[width=0.98\linewidth]{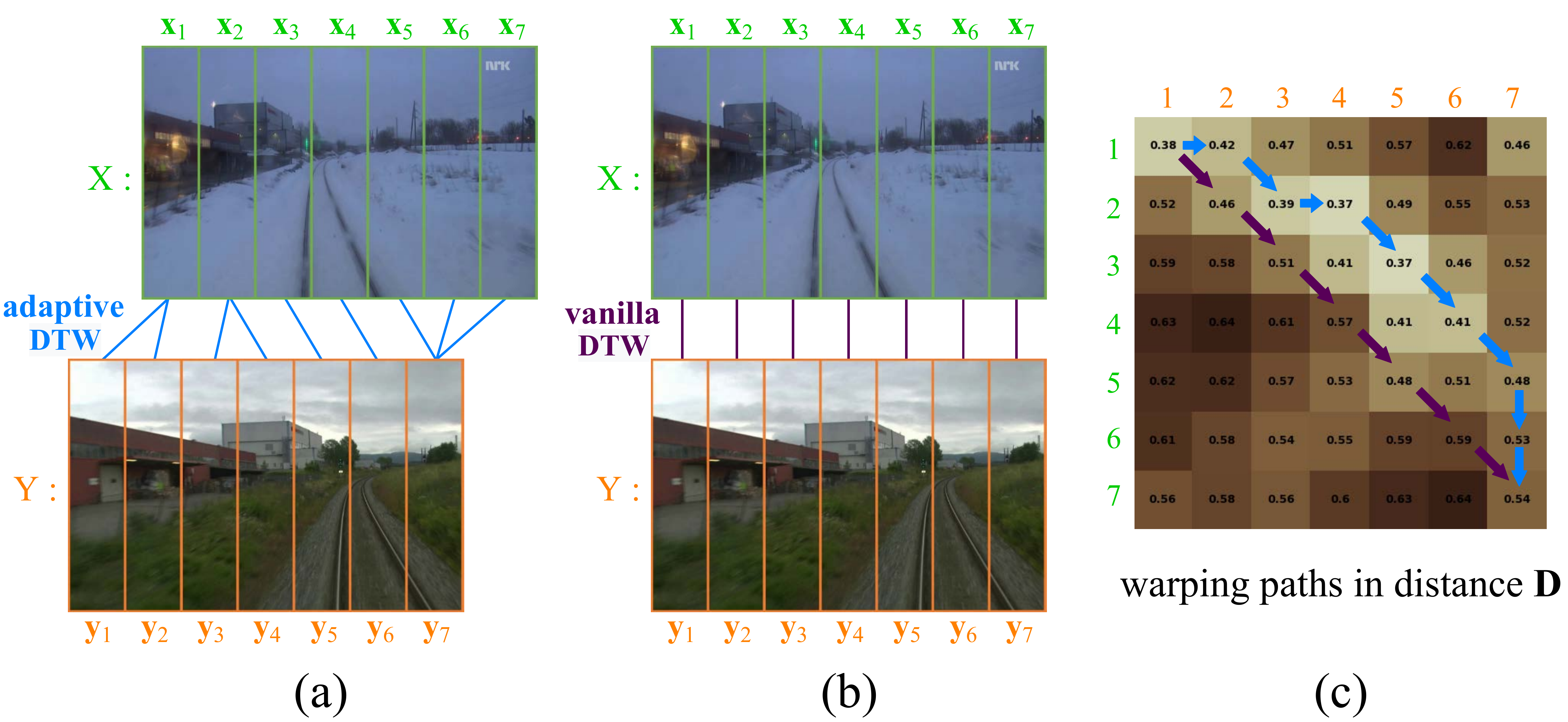}
	\vspace{-0.3cm}
	\caption{
		Example of different spatial alignment ways.
	}
	\vspace{-0.2cm}
	\label{spatial_alignment_compare}
\end{figure}

\begin{table}[!t]
	\caption{F1-Score Comparison of Different Alignment Strategies.}
	\vspace{-0.4cm}	
	\label{table1}
	\begin{center}
		\resizebox{.92\hsize}{!}{$	
			\begin{tabular}{ccccc}
			\toprule
			&\scriptsize{Nordland} &\scriptsize{synth-Nord} &\scriptsize{GP(DL\&DR)} &\scriptsize{GP(DL\&NR)} \\ 
			\midrule
			sliding window & 0.975 & 0.523 & 0.966 &0.955\\
			\midrule
			vanilla DTW & \bf 0.977 & 0.322 & 0.953 &0.947\\
			\midrule
			adaptive DTW & \bf 0.977 & \bf 0.904 & \bf 0.972 & \bf 0.969\\
			\bottomrule
			\end{tabular}
			$}
	\end{center}
	\vspace{-0.8cm}
\end{table}

\begin{table*}[!t]
	\caption{F1-Score Comparison. The Values in ``( )" is the Results of Using Gaussian Random Projection and Restricted Alignment.}
	\vspace{-0.5cm}
	\label{table2}
	\begin{center}
		\begin{tabular}{llllllllllll}
			\toprule
			dataset & SeqSLAM & NetVLAD & SAES  & DenseNet & VGG$_{pool4}$ & VGG$_{fc6}$ & SSM-VPR &  STA-VPR$_{DenseNet}$ & STA-VPR$_{VGG}$ \\
			\midrule
			Nordland   &0.982 &0.587 &0.682 &0.919 &0.987 &0.221 &0.952 &0.977 (0.971) &\textbf{0.994} (\textbf{0.994})     \\
			\midrule
			synth-Nord &0.051 &0.457 &0.072 &0.116 &0.114 &0.098 &0.864 &0.904 (0.863) &\textbf{0.985} (0.979)  \\
			\midrule
			GP(DL\&DR) &0.780 &0.964 &0.649 &0.898 &0.847 &0.916 &\textbf{0.997} &0.972 (0.960) &0.979 (0.978)  \\
			\midrule
			GP(DL\&NR) &0.122 &0.758 &0.367 &0.834 &0.738 &0.671 &0.950 &0.969 (0.947) &\textbf{0.979} (0.953)  \\
			\midrule
			UA       &0.687 &0.574 &0.794 &0.955 &0.979 &0.492 &0.977 &0.968 (0.963) &\textbf{0.991} (0.990)  \\
			\midrule
			Berlin\_A100 &0.457 &0.843 &0.603 &0.714 &0.683 &0.527 &\textbf{0.927} &0.873 (0.857) &0.857 (0.801)  \\
			\bottomrule    
		\end{tabular}
	\end{center}
	\vspace{-0.8cm}
\end{table*}

In this subsection, we conduct experiments to study the effects of different alignment strategies, i.e. compare the adaptive DTW with the vanilla DTW. Besides, we also apply a simple sliding window method (window size is 4) as the baseline. All strategies use temporal alignment, differing only in the way of spatial alignment. The F1-scores of them are shown in Table \ref{table1}. The results show that the performance of adaptive DTW is not worse than that of vanilla DTW. The two strategies get the same F1-score on the Nordland dataset, because spatial alignment does not help when there is no viewpoint change. However, the adaptive DTW achieves better results than the vanilla DTW under varying viewpoints (especially on the synth-Nord dataset). Fig. \ref{spatial_alignment_compare} is an example of different spatial alignment ways, which reveals the reason for misalignment caused by the vanilla DTW. To find the optimal path in the matrix $\mathbf D$, the vanilla DTW chooses a path that goes directly from the upper left to the lower right. This path only passes 7 points, so the total distance is smaller. But it is wrong. However, the adaptive DTW has an adaptive parameter $a>1$ under severe viewpoint changes, which increases the cost of the oblique path, thus leading to correct alignment.

\subsection{Comparison with the state-of-the-art methods}
This subsection evaluates the recognition performance by comparing the F1-scores of the following methods:
\begin{itemize}
	\item\textbf {SeqSLAM} \cite{seqslam}: A classical sequence-based method that can get excellent results under pure appearance changes. For fairness, we set the sequence length ``$ds$" to 20, which is the same thing as $l$ = 20 in our method.
	\item\textbf {NetVLAD} \cite{netvlad}: A viewpoint-robust CNN model for VPR. It can achieve great performance on most datasets.
	\item\textbf {SAES} \cite{MetricLearning}: A deep metric learning VPR model trained on SPED dataset using ``SAES" metric.
	\item\textbf {DenseNet:} It is the ``no alignment"  in subsection \ref{sec_AblationStudy}.
	\item\textbf {VGG$\bm{_{pool4}}$} and \textbf{VGG$\bm{_{fc6}}$}: No alignment methods using $pool4$ and $fc6$ layers of VGG16 pre-trained on ImageNet.
	\item\textbf {SSM-VPR} \cite{SSM1, SSM2}: A highly robust VPR method that is based on the spatial matching of CNN features and achieves near-perfect precision on several datasets.
	\item\textbf {STA-VPR$\bm{_{ DenseNet}}$} and \textbf{STA-VPR$\bm{_{VGG}}$}: Our methods based on DenseNet and VGG$_{pool4}$ respectively. Note that although the feature extracted by VGG$_{pool4}$ is 14$\times$14$\times$512-D, for convenience and fairness, we split it vertically into 7 separate 14336-D local features.
\end{itemize}
For fairness, the NetVLAD and SSM-VPR we used are both based on VGG16. The complete results are shown in Table \ref{table2}. The STA-VPR methods achieve excellent F1-scores on all the datasets we used, while STA-VPR$_{VGG}$ performs better than STA-VPR$_{DenseNet}$ on most datasets. The SeqSLAM and the methods (DenseNet, VGG$_{pool4}$) that use mid-layer deep features can achieve great performance under pure appearance changes (i.e. Nordland). However, they fail on the synth-Nord dataset with severe viewpoint changes. The STA-VPR methods perform well on Nordland and synth-Nord, indicating that they have both great condition invariance and viewpoint invariance. When viewpoint changes dominate (i.e. on GP(DL\&DR)), NetVLAD and VGG$_{fc6}$ achieve better results than DenseNet and VGG$_{pool4}$. The NetVLAD benefits from the viewpoint-invariant NetVLAD pooling layer, while the VGG$_{fc6}$ contains high-level semantic information that is robust to viewpoint changes. Nevertheless, they are all surpassed by STA-VPR utilizing alignment. Moreover, VGG$_{pool4}$ performs better than DenseNet on the Nordland and UA datasets but is weaker on the synth-Nord and GP datasets. We think this is because VGG$_{pool4}$ has encoded more detailed features and less semantic information than DenseNet. This makes it exhibit better condition invariance but weaker viewpoint invariance. However, STA-VPR$_{VGG}$ outperforms STA-VPR$_{DenseNet}$ on all these datasets (except Berlin\_A100). It indicates that richer and more detailed (rather than more semantical) features are better suited to use STA-VPR for higher performance.

Besides, SSM-VPR is a viewpoint-invariant and condition-invariant method, which sets the state of the art in VPR. It also considers the spatial relationship of local features. But its idea is completely different from ours. The focus of SSM-VPR is an exhaustive geometric consistency verification, while that of our method is to dynamically align local features. Our method outperforms SSM-VPR on four datasets, especially on the synth-Nord dataset. However, it is defeated on the two datasets without severe condition changes. This shows that our method is slightly weaker than SSM-VPR when viewpoint changes dominate, but it is more suitable to work under severe variations in condition and viewpoint simultaneously.

\begin{figure}[!t]
	\centering
	\includegraphics[width=0.9\linewidth]{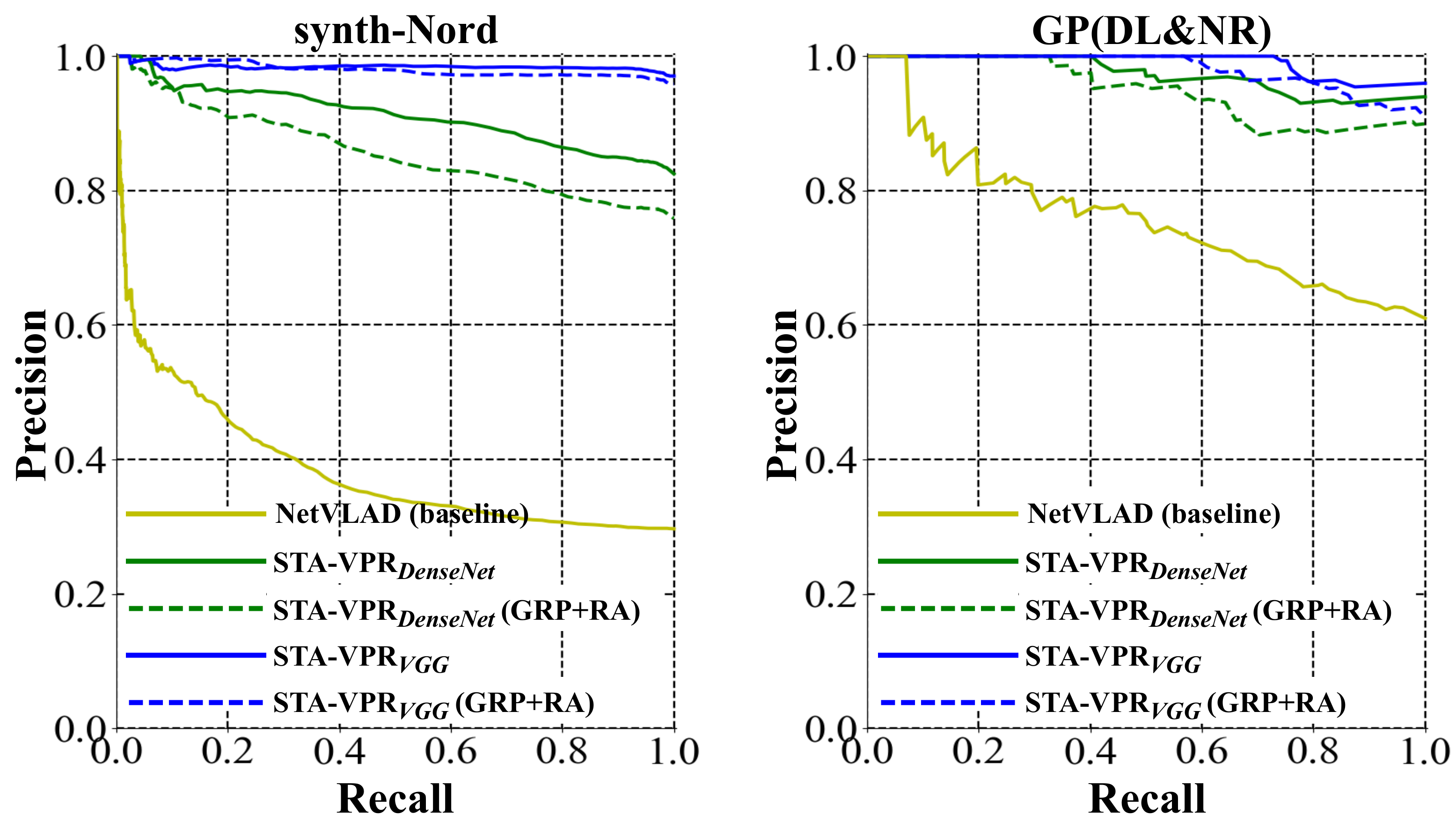}
	\vspace{-0.3cm}
	\caption{
		Precision-recall curves of STA-VPR with and without GRP and RA.
	}
	\vspace{-0.6cm}
	\label{PR}
\end{figure}

From the F1-scores in parentheses in Table \ref{table2}, we can see that the performance degradation is small after using Gaussian Random Projection (GRP) and restricted alignment (RA). Fig. \ref{PR} compares the precision-recall curves on the synth-Nord and GP(DL\&NR) datasets before and after using GRP and RA.

\subsection{Runtime Analysis}
In this subsection, we evaluate the time benefit of GRP and RA. We implement our VPR system in three steps: 1) extract features; 2) compute the distance matrix $\mathbf D_{CH}$ between the current sequence $C$ and the historical sequence $H$ in parallel; 3) take $n$ candidate matrices $\mathbf D_{CT'}$ from $\mathbf D_{CH}$, and then use these candidate $\mathbf D_{CT'}$ to run the LM-DTW process in parallel to find the sequence $T$, we call this step the retrieval process. We use a 20-frame sequence to search the matched sequence in the 1000- and 2000-frame historical sequences for runtime evaluation. The experiments are run on an NVIDIA GeForce GTX Titan X GPU. For a single image, it costs approximately 0.032s to extract the mid-layer features from the DenseNet161 using \emph{pytorch}. The parallel operation is implemented on GPU using the \emph{numba} toolkit in Python. The runtime of computing matrix $\mathbf D_{CH}$ and the retrieval process are shown in Table \ref{tableRuntime}. Obviously, computing $\mathbf D_{CH}$ takes most of the time, which can be accelerated by 30+ times after using GRP and RA.
\begin{table}[!t]
	\caption{The Runtime of Computing the Matrix $D_{CH}$ and the Retrieval Process. ``original" Means No Use of GRP and RA. In Addition, We Use the Runtime of NetVLAD (Excluding the Runtime of Extracting Features) as the Baseline.}
	\vspace{-0.8cm}	
	\label{tableRuntime}
	\begin{center}	
		\begin{tabular}{|c|c|c|c|c|c|c|}
			\hline
			\multirow{2}{*}{task} & \multicolumn{3}{c|}{\bf {computing matrix $\mathbf D_{CH}$}} & \multirow{2}{*}{\bf retrieval} & \multirow{2}{*}{\begin{tabular}[c]{@{}c@{}}NetVLAD\\ (baseline)\end{tabular}}\\ \cline{2-4}
			& original     & GRP        & GRP+RA    &  &                          \\ \hline
			20 vs. 1000       & 0.943s       & 0.038s     & \bf 0.024s    & \bf 0.004s   &  0.285s    \\ \hline
			20 vs. 2000       & 1.976s       & 0.076s     & \bf 0.053s    & \bf 0.005s   &  0.612s    \\ \hline
		\end{tabular}
	\end{center}
	\vspace{-0.8cm}
\end{table}

\section{Conclusion}
Here we proposed the STA-VPR method based on the idea of aligning deep features from both spatial and temporal domains. Specifically, we proposed the adaptive DTW algorithm for spatial alignment and the LM-DTW algorithm for temporal alignment. The adaptive DTW can align local features better than vanilla DTW, while the LM-DTW is insensitive to speed changes and ensures linear time complexity for retrieval. The experimental results showed that our method significantly improves the performance of the method using mid-layer deep features and exhibits strong robustness against variations in appearance and viewpoint simultaneously. It outperformed the state-of-the-art methods on several datasets while showing good speed performance. 
\nocite{*}
\bibliographystyle{IEEEtran}
\bibliography{STA-VPR}

\end{document}